 \documentclass[preprint,5p,times,twocolumn]{elsarticle}

\usepackage{amsmath,amsfonts}
\usepackage{amssymb}
\usepackage{lipsum}
\usepackage{bbding}
\usepackage{makecell}
\usepackage{booktabs} 

\journal{Fundamental Research}

\begin{document}

\begin{frontmatter}



\title{TGRHuman: Text-Guided Realistic 3D Human Generation via Diffusion
Renderer}



\author[label1]{Muxin Zhang}
\author[label2]{Chaohui Yu}
\author[label1]{Yuanwang Yang}
\author[label2]{Min Wei}
\author[label2]{Zhuo Su}
\author[label1]{Kun Li$^{*,}$}

\affiliation[label1]{
    organization={Tianjin University},
    city={Tianjin},
    postcode={300350},
    country={China}
}

\affiliation[label2]{
    organization={Independent Scholar},
    city={Beijing},
    postcode={100080},
    country={China}
}


\begin{abstract}
Realistic 3D human generation plays a crucial role in many graphics applications. However, current methods still struggle to generate high-quality human geometry and texture while maintaining 3D consistency and inference efficiency. In this work, we address these limitations by introducing TGRHuman, a novel approach for generating realistic 3D humans from text. Our method decouples geometry and texture generation to alleviate the issues commonly encountered in NeRF-based methods. Instead of relying on slow, implicit score-distillation-based optimization, we directly use explicit multi-view observation generation and optimization for efficient 3D synthesis. For geometry generation, we propose a high-resolution generative module for multi-view normals together with a geometry-carving strategy that preserves view consistency and supports loose clothing. For texture generation, we produce spatially consistent RGB observations from densely sampled surrounding views using a carefully designed texture-prior acquisition strategy and a diffusion renderer, enabling detailed human texture synthesis. Experiments show that our method can generate high-quality and consistent 3D human geometry and texture efficiently. TGRHuman outperforms existing text-to-3D human methods in both geometry and texture quality.
\end{abstract}




\end{frontmatter}



\begin{figure*}[!ht]
		\centering
		\includegraphics[scale=0.65]{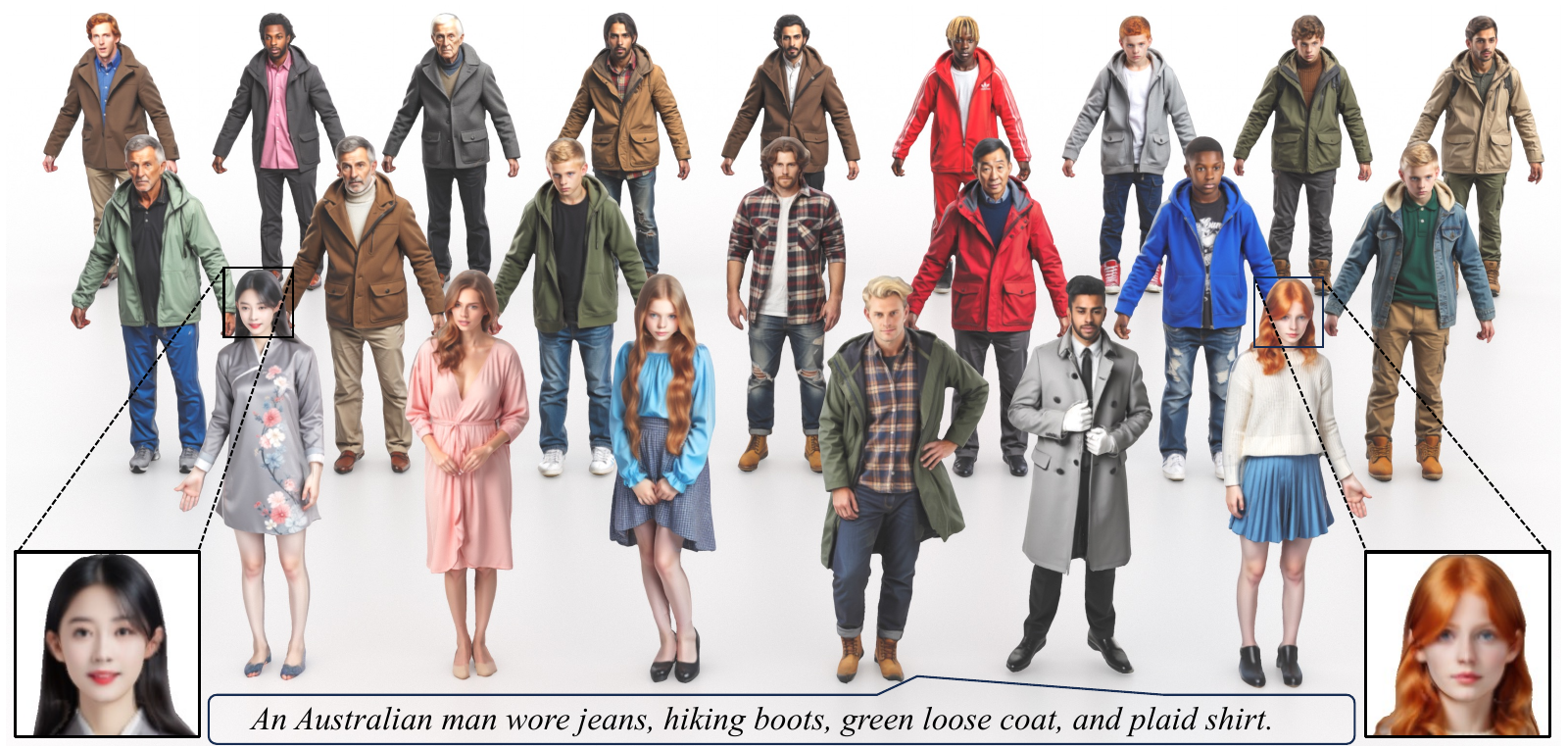}
		\caption{Given text descriptions as input, TGRHuman generates diverse and realistic 3D humans with high-quality geometry and texture efficiently. Our approach not only supports humans wearing loose clothing, but also outputs explicit meshes and texture maps, facilitating downstream graphics applications. The key idea is to leverage consistent 2D multi-view observations together with explicit optimization for efficient 3D human generation.}
		\label{Figure:Teaser}
    \vspace{-0.3cm}
	\end{figure*}
\let\thefootnote\relax\footnotetext{* Corresponding author.}
\section{Introduction}
\label{introduction}
Realistic 3D human generation is essential for digital human modeling in immersive VR/AR experiences, gaming, and the film industry, and benefits a wide range of graphics applications.
Existing methods have made notable progress, but they still struggle to reliably generate high-quality human geometry and texture while ensuring multi-view consistency and inference efficiency.
In this paper, we aim to efficiently generate diverse, high-quality, and consistent 3D human geometry and texture from text descriptions, while also supporting humans with loose clothing.

Existing approaches can be broadly divided into native 3D generation methods and two-stage generation methods.
Native 3D generative models are directly trained on 3D representations such as tri-planes~\cite{chan2022triplanes}, SDFs~\cite{park2019deepsdf}, and FOF~\cite{li2022neurips}. Although these methods naturally possess strong 3D awareness and consistency, the feature dimensionality of the 3D representations used for training is typically much higher than that of 2D images. This discrepancy makes it difficult to leverage pre-trained diffusion models~\cite{rombach2022LDM}, often requiring training from scratch and incurring high computational cost. Furthermore, the limited availability of real 3D human datasets hampers generalization.
Two-stage generation methods~\cite{men2024en3d,huang2023humannorm} are based on pre-trained 2D diffusion models and typically adopt score distillation and differentiable rendering to optimize 3D representations such as NeRF~\cite{mildenhall2020nerf} and DMTet~\cite{shen2021dmtet}. 
The Score Distillation Sampling (SDS) loss~\cite{poole2022dreamfusion} from 2D diffusion is commonly used in the optimization process of two-stage methods~\cite{huang2023humannorm,huang2024tech}, but it is highly time-consuming and may take hours per instance. Because these methods rely on 2D diffusion models, they often lack sufficient 3D structural awareness and suffer from poor view consistency, leading to multi-face artifacts and over-smoothed results~\cite{huang2024tech,richardson2023texture}. In addition, some NeRF-based methods~\cite{kolotouros2023dreamhuman,cao2024dreamavatar} struggle to effectively decouple high-quality geometry and texture from volume-rendering results. Some multi-view-diffusion-based methods~\cite{shi2023mvdream,kim2023chupa} find it difficult to synthesize detailed 3D humans because the generated observations are limited in both viewpoint coverage and resolution. Some works~\cite{SCULPT2024} generate displacements based on the SMPL template mesh; however, due to the fixed topology of SMPL, they fail to support loose clothing.

In summary, existing methods fail to simultaneously achieve diversity, high quality, 3D consistency, efficiency, and support for loose clothing in text-driven human geometry and texture generation. In particular, producing high-quality geometry and high-resolution textures often requires complex optimization procedures and substantial computational resources, making it difficult to balance performance and efficiency.

In this work, we introduce TGRHuman to address these challenges in text-driven realistic 3D human generation while maintaining both efficiency and consistency. To avoid the tight coupling of geometry and texture in NeRF-based methods, we generate geometry and the corresponding texture separately.
For efficiency, we utilize explicit multi-view normal maps and RGB observations to synthesize human geometry and texture through fast optimization. This scheme avoids the slow implicit supervision and optimization processes typical of SDS-based methods. Meanwhile, using multi-view normals and RGB observations allows us to leverage the capabilities of pre-trained 2D diffusion models while maintaining sufficient 3D awareness.

Specifically, for high-quality and diverse geometry generation, we carefully design a strategy to generate consistent four-view normal maps at a resolution of 1024 by leveraging both synthetic and real 3D human datasets. The generated high-resolution normals are then fed into our mesh-carving strategy, which supports humans with loose clothing without requiring any refinement process.

Unlike geometry, texture optimization requires RGB observations from many more viewpoints. To generate densely sampled surrounding RGB observations, we first devise a strategy for obtaining a texture prior. Based on this prior, we design a diffusion renderer capable of high-resolution and consistent rendering from dense views for human texture extraction. The 3D models generated by our method are presented in Fig.~\ref{Figure:Teaser}. Both qualitative and quantitative experiments demonstrate the superiority of our approach.

Our main contributions can be summarized as follows:
\begin{itemize}
\item We propose a strategy that decouples geometry and texture generation via explicit 2D observation generation and dimension elevation, making it more efficient than implicit SDS-based methods while preserving diversity.
\item We propose a high-resolution generative module for multi-view normals together with a geometry-carving strategy that is view-consistent and well suited to humans with loose clothing.
\item We propose a human texture prior acquisition strategy and a diffusion renderer to ensure view consistency and obtain dense, surround-view RGB observations for high-resolution texture generation. 
\end{itemize} 

\section{Related Work}
\subsection{3D Human Geometry Generation}
There are two main approaches to 3D geometry generation: one directly models 3D content using 3D representations and datasets, and the other lifts 2D generative priors to 3D content.
\paragraph{Direct 3D Generative Models} Rodin~\cite{rodin2023} first fits a volumetric neural representation for each training sample and then uses diffusion models to learn and sample the distribution of these 3D instances.
GETAvatar~\cite{zhang2023getavatar} is based on a tri-plane representation, using a GAN to generate geometry and texture tri-planes and employing DMTet to extract an explicit mesh. 
Joint2Human~\cite{J2H2024} learns the distribution of the 3D FOF representation~\cite{li2022neurips}, enabling efficient 3D human geometry generation. 
SCULPT~\cite{SCULPT2024} utilizes a StyleGAN-based generator to learn displacement maps on top of SMPL~\cite{SMPL2015} for clothed human generation.
Shi et al.~\cite{shi2024clothedanimation} further explore generative modeling for diverse clothed 3D human animations.
\paragraph{Lifting 2D Generative Priors to 3D Content}
These methods achieve 3D-aware generation by leveraging 2D diffusion models together with image collections, following the rapid development of diffusion models for 3D generation~\cite{wang2025diffusion3dsurvey}. Most of them~\cite{huang2023humannorm,huang2024tech} rely on optimization-based workflows. They optimize 3D representations (e.g., NeRF~\cite{mildenhall2020nerf} and DMTet~\cite{shen2021dmtet}) under the supervision of 2D diffusion priors. Generalizable human NeRF methods such as EG-HumanNeRF~\cite{wang2026eghumannerf} also exploit human priors for efficient sparse-view human rendering. TeCH~\cite{huang2024tech} and HumanNorm~\cite{huang2023humannorm} produce 3D humans by using pre-trained diffusion models together with Score Distillation Sampling (SDS)~\cite{poole2022dreamfusion} to optimize DMTet. However, the SDS-based optimization process is time-consuming (more than 2 hours) for each instance. Chupa~\cite{kim2023chupa} leverages a dual normal-map generation model and geometry optimization to lift 2D diffusion priors to 3D, but it does not support texture generation. En3D~\cite{men2024en3d} employs a tri-plane-based generator to learn a generalizable 3D representation from 2D synthetic data while using DMTet. TADA~\cite{liao2024tada} creates detailed 3D avatars from text using an optimized SMPL-X model with 3D displacements and texture mapping, enhanced by hierarchical rendering and SDS. AvatarVerse~\cite{zhang2023avatarverse} develops a DensePose-conditioned 2D diffusion model to achieve precise and flexible view-consistency control between 2D and 3D representations. Other works~\cite{cao2024dreamavatar,hong2022avatarclip,jiang2023avatarcraft,huang2023dreamwaltz,kolotouros2023dreamhuman,zeng2023avatarbooth,saito2019pifu} also integrate optimization methods to create avatars.

\subsection{3D Human Texture Generation}
Shert~\cite{zhan2024shert} uses SMPL UV space to complement texture information, but it is limited by the topology of SMPL.
HumanNorm and TeCH~\cite{huang2023humannorm,huang2024tech} utilize optimization schemes based on score distillation, which tend to produce overly smooth appearances. Some methods~\cite{albahar2023sgd,ho2024sith,saito2019pifu} rely on back-view estimation and blending.
TEXTure~\cite{richardson2023texture} and Text2Tex~\cite{chen2023text2tex} employ 2D diffusion models to iteratively paint the mesh from each viewpoint, with each step conditioned on previous results. However, such a strategy falls short of capturing global information, resulting in inconsistencies across views. TexFusion~\cite{cao2023texfusion} proposes texture aggregation from different camera views and maintains a latent texture map during denoising. Works such as~\cite{huo2025texgen,perla2024easitex} further improve the texture-sampling strategy to reduce view discrepancy. 
Paint3D~\cite{zeng2024paint3d} introduces a coarse-to-fine generative framework that integrates texture sampling from each new viewpoint with UV inpainting to refine the texture map. However, view inconsistency and local artifacts remain unavoidable due to the limited implicit guidance provided by 2D diffusion priors. 
\subsection{Multi-view Diffusion Model}
Another line of work combines multi-view consistent image generation with 3D reconstruction for 3D content creation. Diffusion models have demonstrated strong multi-view generation ability: MVDream~\cite{shi2023mvdream} simultaneously generates all images with global awareness through cross-view interactions, while Zero-1-to-3~\cite{liu2023zero123} and Wonder3D~\cite{long2024wonder3d} offer zero-shot, viewpoint-conditioned novel-view synthesis from a single image. SV3D~\cite{voleti2025sv3d} utilizes a video diffusion model for multi-view synthesis. For multi-view human image synthesis, MvHuman~\cite{jiang2023mvhuman} proposes a multi-view sampling strategy to generate multi-view images, which are then used to train a neural radiance field for free-view rendering. MagicMan~\cite{he2024magicman} employs a 3D-aware diffusion model and a post-processing procedure to generate multi-view human images from a reference image. More recent methods~\cite{xue2024human3diffusion,li2024pshuman,huang2024mvadapter,xu2023seeavatar,qiu2025LHM,xiang2025trellis2,yang2025humanlift} achieve multi-view generation and reconstruction in both realistic and cartoon styles.

General 3D generation methods are difficult to apply directly to realistic human generation because of their limited output resolution. For example, \textit{MagicMan} and \textit{PSHuman} generate only 24 multi-view images at a resolution of 512 and 6 images at a resolution of 768, which constrains high-quality 3D human generation. Rather than relying solely on unstable and complex multi-view attention, we exploit both novel-view and global cues from coarse vertex colors to generate high-resolution (1024) images from arbitrary viewpoints through our texture prior and diffusion renderer. As summarized in Tab.~\ref{Table:Compare}, our approach can generate high-quality geometry and texture without relying on SDS, enabling efficient generation of realistic and consistent 3D humans.

\begin{table}[!t]
  \small
  \caption{Comparison with existing methods. }
  \label{Table:Compare}
  \centering
  \resizebox{0.47\textwidth}{!}{
  \begin{tabular}{lccccc}
    \toprule
    Method & \makecell[c]{Geometry\\ Quality} &
    \makecell[c]{Texture\\ Quality}&
    \makecell[c]{Realistic} &
    \makecell[c]{Consistency} &
    \makecell[c]{Free of SDS}\\
    \midrule
     Chupa~\cite{kim2023chupa} &\Checkmark & None & \Checkmark & \XSolidBrush &\Checkmark \\   
    TEXTure~\cite{richardson2023texture} & None & \XSolidBrush &  \Checkmark  & \XSolidBrush &\XSolidBrush\\
    TADA~\cite{liao2024tada} & \XSolidBrush & \Checkmark & \XSolidBrush & \Checkmark  &\XSolidBrush\\
    Joint2Human~\cite{J2H2024} & \Checkmark  & None & \Checkmark  & \Checkmark&\Checkmark \\
     SCULPT~\cite{SCULPT2024}& \XSolidBrush & \XSolidBrush & \Checkmark & \Checkmark &\Checkmark\\
    HumanNorm~\cite{huang2023humannorm} &\XSolidBrush & \Checkmark &\XSolidBrush &  \Checkmark &\XSolidBrush \\
     En3D~\cite{men2024en3d} &\XSolidBrush & \Checkmark &\Checkmark &  \XSolidBrush &\Checkmark \\
    \textbf{Ours} & \Checkmark & \Checkmark & \Checkmark & \Checkmark &\Checkmark\\
    \bottomrule
  \end{tabular}
  }
\vspace{-0.5cm}
\end{table}

\section{Methodology}
\begin{figure*}[!ht]
		\centering
		\includegraphics[scale=0.25]{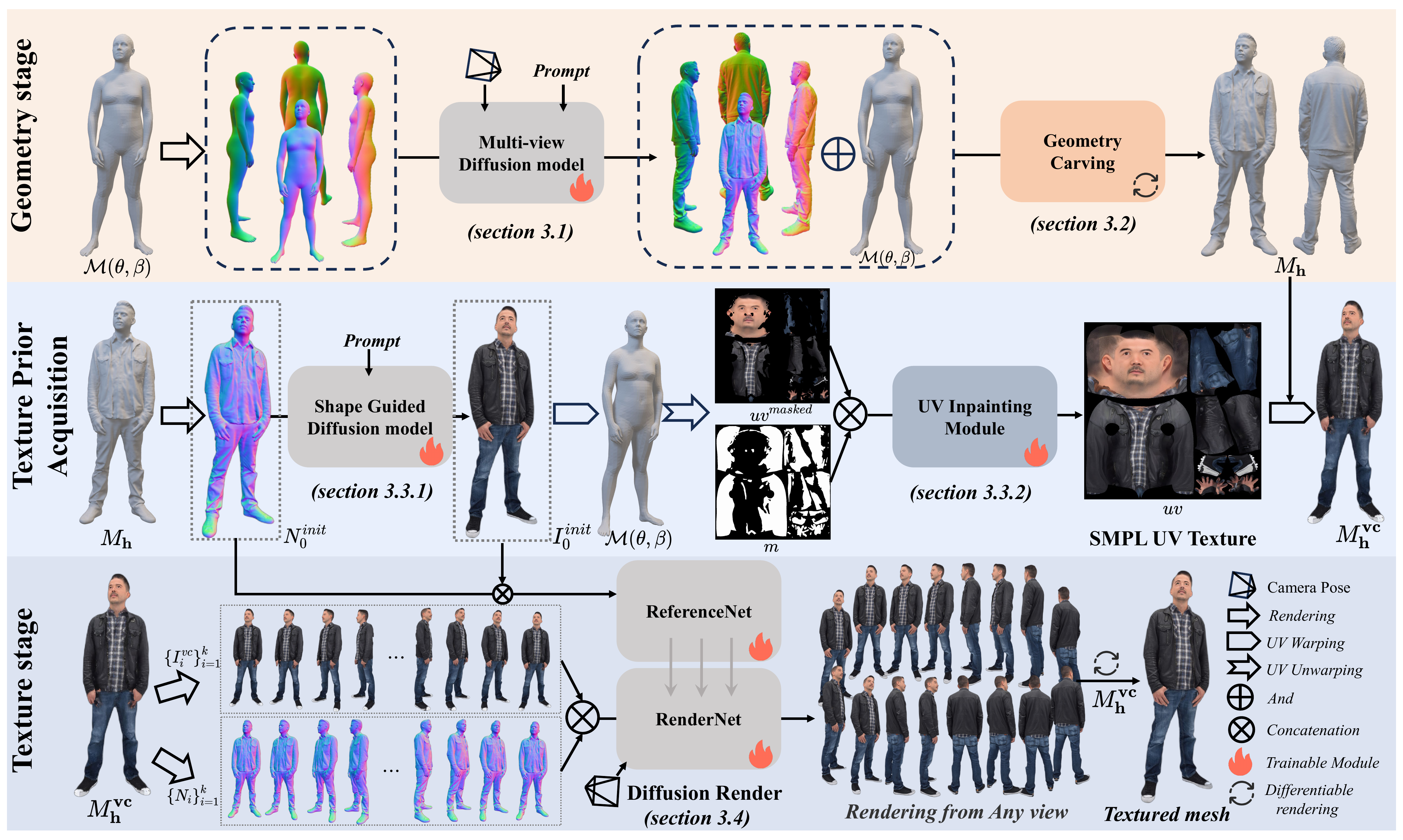}
		\caption{The overall architecture of TGRHuman. We decouple geometry and texture generation to produce realistic, high-quality humans via explicit 2D observation generation and dimension elevation. Specifically, our model includes: (a) a geometry stage with a high-resolution generative module for multi-view normals and a geometry-carving strategy for human shape; (b) a texture-prior acquisition stage with a shape-guided diffusion model for front-view appearance generation and a UV inpainting module for global texture prior construction; and (c) a texture stage with a diffusion renderer based on the texture prior for consistent, dense free-view rendering.}
		\label{fig_net}
    \vspace{-0.3cm}
	\end{figure*}
Our method aims to generate realistic clothed 3D humans with high-quality geometry and texture from text descriptions. To avoid the coupling of texture and geometry in previous methods~\cite{kolotouros2023dreamhuman,cao2024dreamavatar}, we decouple geometry and texture generation via explicit 2D observation generation and dimension elevation. As illustrated in Fig.~\ref{fig_net}, TGRHuman mainly contains two stages. The geometry stage focuses on a high-resolution normal-map generative module (Sec.~\ref{MNG}) and a geometry-carving strategy (Sec.~\ref{geometry_rec}) for human shape reconstruction. In the texture stage, unlike multi-view generation methods with a limited number of views, our approach enables free-view rendering directly through the texture prior (Sec.~\ref{tex_prior}) and diffusion renderer (Sec.~\ref{Diffusion_Renderer}), without training intermediate 3D representations or performing post-processing.
\subsection{Multi-view Human Normals Generation}
\label{MNG}
To capture geometric details globally, we generate normal maps of a clothed human from multiple views. Due to VRAM limitations, previous methods~\cite{kim2023chupa} found it difficult to simultaneously satisfy the requirements on viewpoint number and image resolution for high-quality generation. With our carefully designed pipeline, our method supports the generation of four-view (front, back, left, and right) normal maps at a resolution of 1024. Benefiting from these high-quality geometric observations, our model does not require local post-refinement or super-resolution, unlike Chupa~\cite{kim2023chupa}.

To achieve pose-controllable human generation, we leverage SMPL~\cite{SMPL2015}, a parametric human template denoted as $\mathcal{M}\left(\theta,\beta\right)$, where $\theta$ and $\beta$ represent pose and shape parameters, respectively. We first obtain the SMPL mesh $\mathcal{M}\left(\theta,\beta\right)$ by sampling $\theta$ and $\beta$ to provide pose guidance. We then render the SMPL mesh from four target views as conditional images. The SMPL condition also alleviates self-occlusion issues of the limbs. Next, we employ a pre-trained variational autoencoder (VAE) encoder to transform these SMPL renderings into latent vectors. To control human geometric details, we extract the text condition using CLIP~\cite{radford2021clip}. We encode the camera parameters with MLPs and add the resulting camera embeddings to the time embeddings as residuals, following~\cite{shi2023mvdream}. Conditioned on the SMPL latents, text embeddings, and camera embeddings, we leverage a pre-trained latent diffusion model (LDM)~\cite{rombach2022LDM} to generate normal maps while inheriting the 2D priors of the LDM. The latent vectors from SMPL renderings are concatenated with noise and fed into the denoising UNet.

A common challenge in multi-view generation is maintaining view consistency. To address this issue, we employ cross-view interaction during the diffusion process. This approach enables information from each viewpoint to extend beyond its features, integrating it into a comprehensive representation that includes global details from all views. Specifically, we collect all intermediate results to serve as queries and values when computation reaches a self-attention layer rather than relying solely on the outcomes from the current branch. This strategy ensures effective information integration across different views, guaranteeing coherence in the generated results among these views.

We use the $v$-prediction target~\cite{salimans2022vprediction} during training. The corresponding optimization objective is:
\begin{equation}    \mathcal{L}^{\mathbf{mvn}}=\mathbb{E}_{\mathbf{n},\mathbf{v},\mathbf{y}_{smpl},t,c,\pi}\left[\left\|\hat{\mathbf{v}}_{\theta}\left(\mathbf{n}_t,\mathbf{y}_{smpl},t,c,\pi\right)-\mathbf{v}_t^\mathbf{x}\right\|_2^2 \right],
\end{equation}
where $\mathbf{y}_{smpl}$ denotes the conditional SMPL latents, $t$ denotes the timestep, $c$ is the prompt, and $\pi$ denotes the camera parameters. $\mathbf{n}_t=\alpha_t\mathbf{n}+\sigma_t{\epsilon}_\mathbf{n}$ denotes the latent representation of the human normal maps, where ${\epsilon}_\mathbf{n}\sim\mathcal{N}({0},\mathbf{I})$ is independently sampled noise. $\mathbf{v}_t^\mathbf{n}=\alpha_t\mathbf{\epsilon}_\mathbf{n}-\sigma_t\mathbf{n}$ denotes the $v$-prediction target at timestep $t$. $\sigma_t$ and $\alpha_t$ are the parameters of the diffusion scheduler. To implement classifier-free guidance (CFG), we use blank text embeddings with a probability of 10\% during training. During CFG inference, the model output is extrapolated toward ${\mathbf{v}}_{\theta}\left(\mathbf{n}_t,\mathbf{y}_{smpl},t,c,\pi\right)$ and away from ${\mathbf{v}}_{\theta}\left(\mathbf{n}_t,\mathbf{y}_{smpl},t,\pi\right)$.

\begin{figure}[!t]
		\centering
		\includegraphics[scale=0.6]{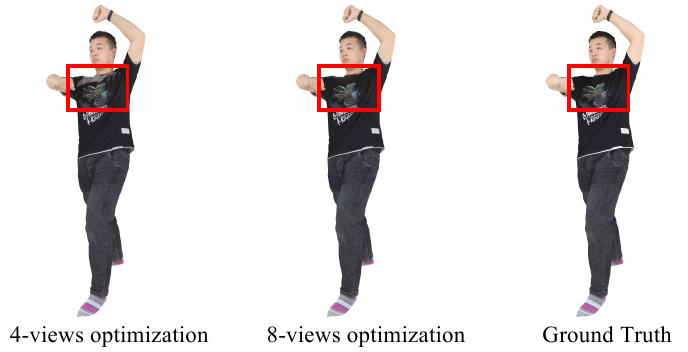}
		\vspace{-0.3cm}
		\caption{Texture optimization results using RGB observations from different numbers of views.}
		\label{fig_net_con}
  \vspace{-0.5cm}
	\end{figure}
    
\subsection{Human Geometry Carving with Normal Maps}
\label{geometry_rec}
In Sec.~\ref{MNG}, we obtain multi-view normal maps $\left(\mathbf{n}^{\mathbf{f}},\mathbf{n}^{\mathbf{b}},\mathbf{n}^{\mathbf{l}},\mathbf{n}^{\mathbf{r}}\right)$ aligned with world coordinates from four views covering 360°. To fuse these observations into a human mesh, we deform the vertices and topology of the initial SMPL mesh $\mathcal{M}\left(\theta,\beta\right)$ into a detailed human mesh $M_{h}$ by comparing the normal maps with renderings produced by a differentiable rasterizer~\cite{Laine2020diffrast}. 
Notably, although our geometry optimization is initialized from SMPL, it still performs well for humans wearing loose-fitting clothing because we dynamically adjust the topology while optimizing vertex displacements. Our goal is to minimize the pixel-aligned error between the normal observations and the rendered results. The optimization loss is formulated as follows:
\begin{equation}
    \mathcal{L}^{\mathbf{rec}} = \mathcal{L}^{\mathbf{n}} + \mathcal{L}^{\mathbf{mask}} + \lambda \cdot\mathcal{L}^{\mathbf{reg}},
\end{equation}
where $\mathcal{L}^{\mathbf{n}} = \sum_{i\in\{\mathbf{f},\mathbf{b},\mathbf{l},\mathbf{r}\}} \|\mathbf{n}^\mathbf{i}-\hat{\mathbf{n}}^\mathbf{i}\|_1$ denotes the normal-rendering loss, and $\mathcal{L}^{\mathbf{mask}} = \sum_{i\in\{\mathbf{f},\mathbf{b},\mathbf{l},\mathbf{r}\}} \|\mathbf{m}^\mathbf{i}-\hat{\mathbf{m}}^\mathbf{i}\|_1$ denotes the mask loss, which characterizes the difference in mesh contours. $\mathcal{L}^{\mathbf{reg}}$ is a regularization term that minimizes the cosine similarity between face normals of neighboring surfaces to encourage smoothness, and $\lambda=1$. After each optimization step, following~\cite{palfinger2022continuous}, we remesh the intermediate geometry to obtain a new topology by merging or splitting triangle faces. In this way, by leveraging the strong geometric prior provided by the SMPL mesh, we transform the generated 2D normal maps into a 3D human mesh. We can further adopt Poisson reconstruction to enhance visual quality, or use the SMPL-H hand model to replace the generated hand geometry as in~\cite{xiu2023econ}. Other than that, our human geometry does not require additional post-processing.
\subsection{Human Texture Prior Acquisition from SMPL}
\label{tex_prior}
In Sec.~\ref{geometry_rec}, by leveraging the SMPL mesh as a robust geometric prior, we can optimize human geometry using normal maps from only four views. However, because no texture prior is available, directly applying the same mechanism to texture generation is highly challenging. As shown in Fig.~\ref{fig_net_con}, four-view RGB images fail to cover certain self-occluded regions. In addition, generating a large number of consistent RGB observations from multiple views simultaneously is computationally demanding. In this section, we propose a strategy to construct a robust texture prior based on SMPL UV space. We first apply a shape-aligned diffusion model to generate a coarse front-view appearance aligned with the geometry. We then unwrap this front-view observation into SMPL UV space and use a UV inpainting model to complete the missing regions.
\paragraph{Front Appearance Generation Aligned with the Shape} 
To initialize and guide the diffusion renderer, we first select a view of the current geometry and obtain the initial texture observation $I_{0}^{init}$ from camera view $p_{0}$. We use the normal map rendered from this viewpoint as a condition to ensure that the initial texture aligns with the human geometry. We formulate this step as a Pix2Pix task and progressively fine-tune a shape-guided diffusion model on both synthetic and real datasets. During training, we extract normal-map latents using the VAE and concatenate them with noise to train the UNet of the LDM.
As in Sec.~\ref{MNG}, we also use the $v$-prediction objective. The training objective of the shape-guided diffusion model is:
\begin{equation}
\mathcal{L}^{\mathbf{init}}=\mathbb{E}_{\mathbf{x},\mathbf{v},\mathbf{y}_{normal},t,c}\left[\left\|\hat{\mathbf{v}}_{\theta}\left(\mathbf{x}_t,\mathbf{y}_{normal},t,c\right)-\mathbf{v}_t^\mathbf{x}\right\|_2^2 \right],
\end{equation}
where $\mathbf{y}_{normal}$ corresponds to the rendered normal maps.
\paragraph{SMPL UV Unwrapping and Completion} We aim to extract a complete texture map of the SMPL template from the generated front-view observation so as to capture the full texture and ensure consistency. First, we extract the visible portion of the SMPL texture map by leveraging the correspondence among pixels in the front-view image, faces of both the human mesh and the SMPL mesh, and UV texture pixels, as shown in Fig.~\ref{fig_net}. 
Next, we compute the visibility of each face and the SMPL UV mask via ray casting. In this way, we project the visible appearance and mask from image space to UV space. To complete the invisible regions of the UV texture, we fine-tune the Stable Diffusion inpainting model~\cite{rombach2022LDM}. The optimization objective is:
\begin{equation}    \mathcal{L}^{\mathbf{inpainting}}=\mathbb{E}_{uv^{masked},\mathbf{v},m,t,c,}\left[\left\|\hat{\mathbf{v}}_{\theta}\left(uv^{masked},m,t,c\right)-\mathbf{v}_t^\mathbf{x}\right\|_2^2 \right],
\end{equation}
where $uv^{masked}$ denotes the masked UV map, $m$ denotes the UV mask, $t$ denotes the timestep, and $c$ is the prompt.
During inference, we feed the partial UV texture map $uv^{masked}$ and the corresponding visibility mask $m$ into the UV inpainting module to obtain a completed SMPL UV texture map $uv$ for the SMPL mesh $\mathcal{M}\left(\theta,\beta\right)$.

\paragraph{SMPL UV Mapping to Human Geometry}
Given the texture map of SMPL, we project it onto the human mesh $M_{\mathbf{h}}$ through vertex coloring. This process results in the initial human textured mesh $M_{\mathbf{h}}^{\mathbf{vc}}$, which serves as an approximation of the final human texture. In this way, we effectively establish the human texture prior.
\begin{figure*}[h]
  \centering
   \includegraphics[width=1.02\textwidth]{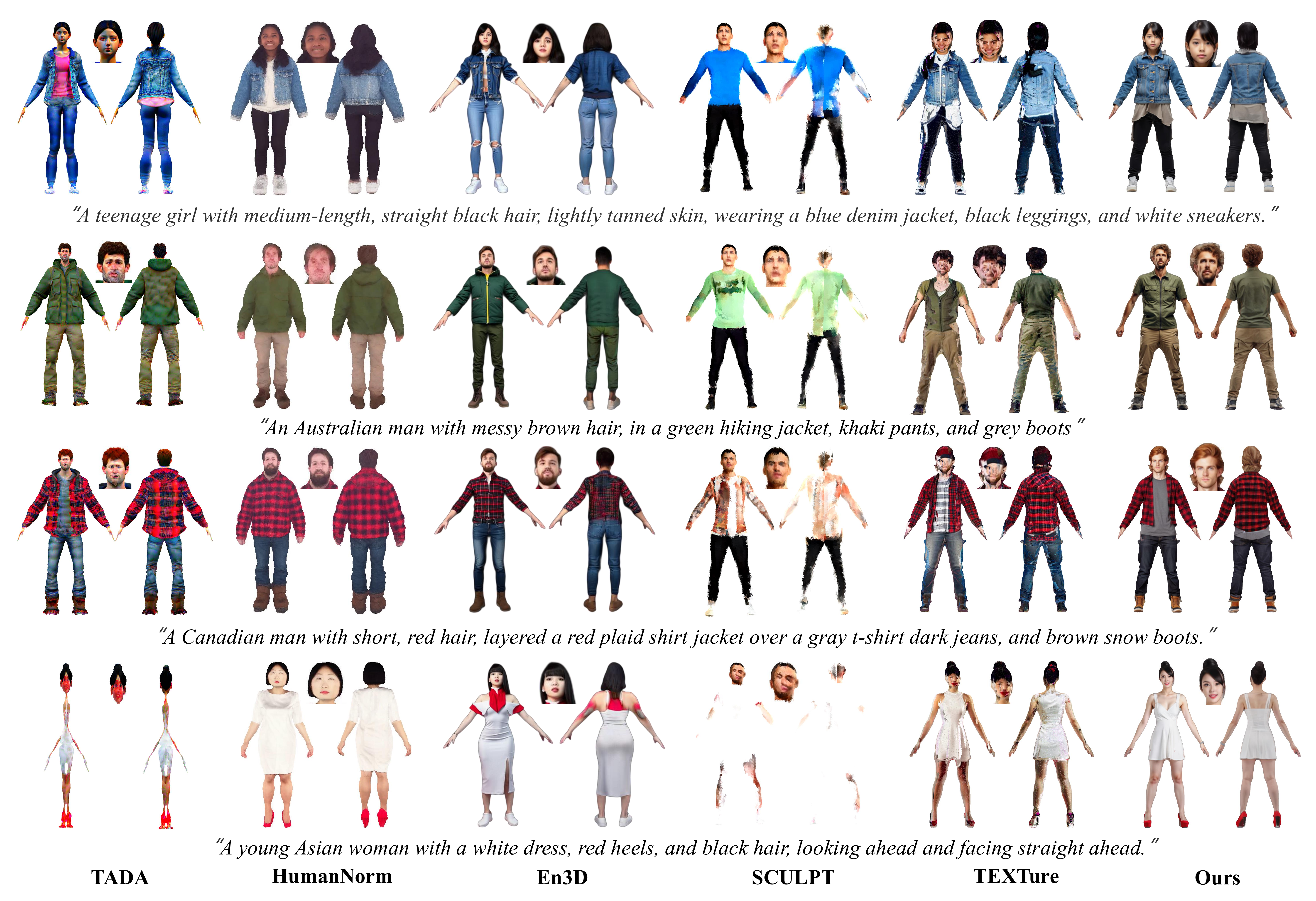}
   \caption{Qualitative comparison of human textures. For fair comparison, our method does not perform post-processing during inference, such as replacing hands with the SMPL model. Our approach generates highly realistic appearances.}
    \label{fig:compare_texture}
    \vspace{-0.2cm}
\end{figure*}

\begin{figure*}[h]
  \centering
   \includegraphics[width=1.0\textwidth]{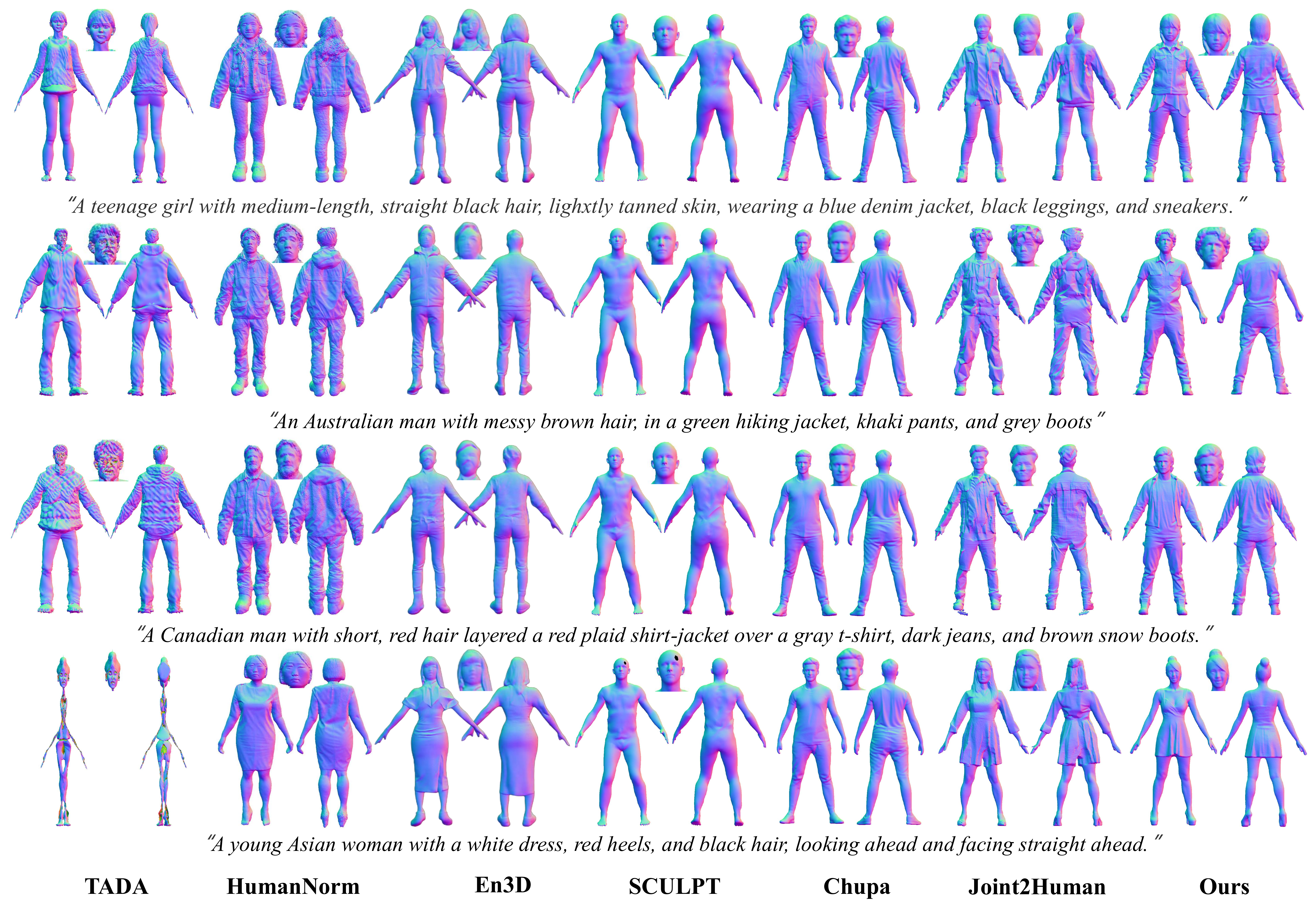}
   \caption{Qualitative comparison of human geometry. The normal maps are rendered from the generated 3D human meshes.}
    \label{fig:compare_geometry}
\end{figure*}
\subsection{Texture Painting with Diffusion Renderer}
\label{Diffusion_Renderer}
In this stage, we aim to obtain realistic texture for the human geometry generated in Sec.~\ref{geometry_rec}. Some methods~\cite{voleti2025sv3d,jiang2023mvhuman} create textures by generating multi-view images, similar to multi-view normal generation. However, such a scheme struggles to recover accurate textures for self-occluded regions of the human mesh because the observation views are sparse and low-resolution. To address this issue, based on the texture prior proposed in Sec.~\ref{tex_prior}, we introduce a 360-degree high-resolution diffusion renderer to obtain a UV texture map for the mesh.
\paragraph{Free-View Rendering with the Diffusion Renderer} To achieve 360-degree rendering, we propose a reference-based rendering strategy and use a diffusion model as the renderer.
Given camera views $\left \{ p_{i} \right \}_{i=1}^{k}$, we first render $M_{\mathbf{h}}^{\mathbf{vc}}$ to obtain RGB images $I_{i}^{vc}$ and normal maps $N_{i}$ as guidance conditions. The normal maps encode geometric details and mesh orientation, while the RGB observations of $M_{\mathbf{h}}^{\mathbf{vc}}$ provide important texture priors. Using these conditions, we train our diffusion renderer.

\begin{figure}[!h]
		\centering
		\includegraphics[scale=0.23]{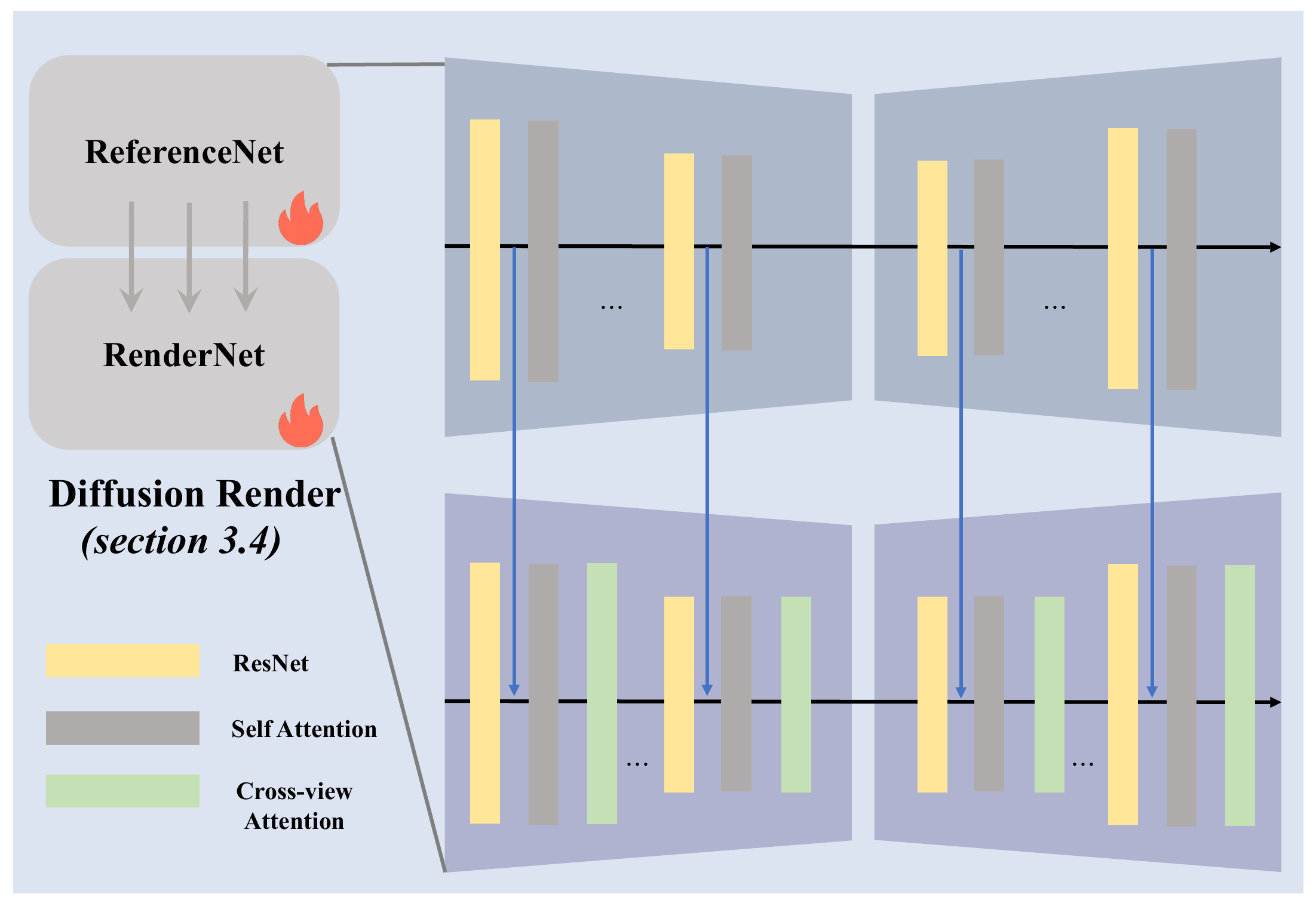}
		\vspace{-0.3cm}
		\caption{The detailed architecture of the diffusion renderer.}
		\label{fig_diff_render}
  \vspace{-0.5cm}
	\end{figure}

Our diffusion renderer consists of two modules: the main RenderNet and a ReferenceNet. The input to RenderNet includes the camera parameters $\pi_{i}$ corresponding to view $p_{i}$, the rendered RGB image $I_{i}^{vc}$, and the normal map $N_{i}$ of the initial textured human mesh $M_{\mathbf{h}}^{\mathbf{vc}}$. These inputs are first encoded by the VAE into image latents and then concatenated with the latent noise.
To ensure multi-view consistency, we inject ReferenceNet features into the frozen RenderNet layer by layer. The appearance and geometry features of the same character, namely $I_{0}^{init}$ and $N_{0}^{init}$, are fed into ReferenceNet, which is designed to maintain identity consistency between renderings from different views and the front observation $I_{0}^{init}$. Specifically, as shown in Fig.~\ref{fig_diff_render}, the core idea is to sum the self-attention input features from ReferenceNet and RenderNet, and then feed the combined feature into the self-attention layers of RenderNet's UNet. This design allows RenderNet to directly reference features from ReferenceNet during self-attention computation, thereby improving the 3D consistency of appearance and texture.

Both RenderNet and ReferenceNet are initialized from clones of a pre-trained latent diffusion model~\cite{rombach2022LDM}. The difference is that the former uses cross-view attention layers, whereas the latter does not. Our training process is divided into two stages. In Stage 1, we train RenderNet with the following objective:
\begin{equation}    
\mathcal{L}^{\mathbf{render}}=\mathbb{E}_{\mathbf{I},\mathbf{v},I^{vc},N,t,\pi}\left[\left\|\hat{\mathbf{v}}_{\theta}\left(\mathbf{I}_{t},I^{vc}, N, t,\pi\right)-\mathbf{v}_t^\mathbf{x}\right\|_2^2 \right],
\end{equation}
where $I^{vc}$ and $N$ denote the texture and geometry guidance conditions derived from $M_{\mathbf{h}}^{\mathbf{vc}}$.

In Stage 2, the parameters of RenderNet $\theta$ are frozen, and only the parameters of ReferenceNet $\phi$ are optimized via $\mathcal{L}^{\mathbf{ref}}$, while keeping the same $v$-prediction objective as in $\mathcal{L}^{\mathbf{render}}$. The optimization objective for ReferenceNet is:
\begin{equation}
\mathcal{L}^{\mathbf{ref}}=
\mathbb{E}_{\mathbf{I},\mathbf{v},I^{vc},N,
I_{0}^{init},N_{0}^{init},t,\pi}
\left[
\left\|
\hat{\mathbf{v}}_{\theta,\phi} 
\big( 
\mathbf{I}_{t},
I^{vc}, N,
I_{0}^{init}, N_{0}^{init},
t,\pi
\big)
-
\mathbf{v}_t^\mathbf{x}
\right\|_2^2
\right],
\end{equation}

During inference, we first sample $k=32$ successive viewpoints around the yaw axis of the human and render the corresponding normals $\left \{ N_{i} \right \}_{i=1}^{k}$ and texture priors $\left \{I_{i}^{vc}\right \}_{i=1}^{k}$. Guided by the front observation $I_{0}^{init}$, RenderNet produces realistic observations $\left \{I_{i}^{real}\right \}_{i=1}^{k}$ from arbitrary selected views. From these rendered observations, we can easily extract a complete texture map for the human mesh while avoiding the occlusion issues caused by sparse views.\\
\vspace{-0.3cm}
\paragraph{Multi-view Rendering Integration for the Texture Map}
Given the multi-view rendering observations and the human mesh $M_{\mathbf{h}}$, we aim to recover the texture map of $M_{\mathbf{h}}$. First, we use XAtlas~\cite{xatlas} to generate UV coordinates for each vertex of $M_{\mathbf{h}}$. Based on the initial textured human mesh $M_{\mathbf{h}}^{\mathbf{vc}}$ and the UV coordinates produced by XAtlas, we obtain a coarse texture map $\hat{T}$ through UV unwrapping. Then, using the paired data $\left \{ \pi_{i},I_{i}^{real} \right \}_{i=1}^{k}$, we optimize the texture map $\hat{T}$ with a differentiable rasterizer~\cite{Laine2020diffrast}. The optimization objective is:
\begin{equation}
\mathcal{L}^{\mathbf{tex}} = 
\mathcal{L}^{\mathbf{T}} 
+ {\lambda}_{\mathrm{ssim}} \cdot\mathcal{L}^{\mathbf{ssim}} 
+ {\lambda}_{tv} \cdot\mathcal{L}^{\mathbf{tv}},
\end{equation}
where $\lambda_{ssim}=10$ and $\lambda_{tv}=1$. $\mathcal{L}^{\mathbf{T}}$ denotes the L1 loss between the rendered RGB image and the observation $I_{i}^{real}$, $\mathcal{L}^\mathbf{ssim}$ is the SSIM loss~\cite{wang2004ssim} that measures structural similarity between two images, and $\mathcal{L}^{\mathbf{tv}}$ denotes the total variation loss~\cite{rudin1992nonlinear}, which encourages smoothness while preserving details.

\section{Experiments}
\subsection{Experimental Setup}

\noindent \textbf{Training data.} We train TGRHuman sequentially on synthetic and real human datasets. For real human data, we use 10k human scans from the THuman2.1~\cite{tao2021function4d}, 2K2K~\cite{han2023high}, and Human4DiT~\cite{shao2024human4dit} datasets, and allocate 50 samples from THuman2.0, 200 from THuman2.1, 200 from 2K2K, and 500 from Human4DiT for evaluation. The RGB and normal images are rendered at a resolution of 1024 using both perspective and orthographic cameras from 32 fixed views. These viewpoints are evenly distributed with azimuth angles ranging from 0 to 360 degrees, ensuring comprehensive coverage of each scan.\\
\textbf{Baselines.} 
We compare our 3D synthesis results with the text-to-3D human SOTA methods, including HumanNorm, En3D, TADA, Joint2Human, Chupa, SCULPT, and TEXTure. Additionally, to demonstrate the superiority of our proposed diffusion renderer, we compare our method with several novel view synthesis works, including Wonder3D, SV3D, Zero123, and the latest method, MagicMan, for human-specific novel view synthesis.\\
\textbf{Metrics.}
Evaluation is conducted on two tasks: (1) human texture and geometry generation, where we use FID (Fréchet Inception Distance) to evaluate the quality of generated textures and geometry, and then use the CLIP score to assess the compatibility between prompts and rendered views of the generated 3D human models; and (2) novel-view synthesis in the texture stage, where we compare generated views with ground-truth images using PSNR, SSIM~\cite{wang2004ssim}, and LPIPS~\cite{zhang2018unreasonable}. 

\begin{table}[!h]
\centering
\caption{Quantitative evaluation on the 3D human generation.}
\label{Tab_nvs}
\setlength{\tabcolsep}{2pt}
\begin{tabular}{lcccc}
\toprule
 Methods & FID$_{normal}\downarrow$ & FID$_{rgb}\downarrow$ & CLIP$_{rgb}\uparrow$ &CLIP$_{normal}\uparrow$\\
\hline
Chupa & 32.91  \quad  & -  & -  & 0.0816\\
TEXTure & -  \quad  & 53.56  & 0.2318& - \\
TADA & 47.56  \quad  & 40.74  & 0.2297 & 0.1419\\
Joint2Human& 31.24  \quad  & - & - & 0.1903 \\
SCULPT & 55.79  \quad  & 50.91  & 0.1306& 0.0986\\
HumanNorm & 40.14  \quad  & 34.72  & 0.2547& 0.1840\\
En3D & 33.86  \quad  & 28.64 & 0.2334 & 0.1589\\
Ours & \textbf{29.48 } \quad  &\textbf{ 25.36}  & \textbf{0.2552} & \textbf{0.2171}\\
\bottomrule
\end{tabular}
\end{table}

\begin{table}[!h]
\centering
\caption{Quantitative evaluation on 3D human generation under complex poses.}
\label{Tab_nvs_com_pose}
\setlength{\tabcolsep}{2pt}
\begin{tabular}{lcccc}
\toprule
 Methods & FID$_{normal}\downarrow$ & FID$_{rgb}\downarrow$ & CLIP$_{rgb}\uparrow$ &CLIP$_{normal}\uparrow$\\
\hline
Chupa & 33.88  \quad  & -  & -  & 0.0750\\
Joint2Human& 36.72  \quad  & - & - & 0.1508 \\
Ours & \textbf{28.91 } \quad  & -  & - & \textbf{0.2059}\\
\bottomrule
\end{tabular}
\end{table}

\begin{table}[!h]
\centering
\caption{Quantitative evaluation on the novel view synthesis.}
\label{Tab_human}
\setlength{\tabcolsep}{10pt}
\begin{tabular}{lccc}
\toprule
 Methods & PSNR $\uparrow$ &  SSIM $\uparrow$ & LPIPS $\downarrow$ \\
\hline
MagicMan & 22.4  \quad  & 0.937  & 0.051  \\
Wonder3D& 15.9  \quad  & 0.895  & 0.106  \\
SV3D & 12.2  \quad  & 0.798 & 0.207\\
Stable Zero123 & 16.1  \quad  & 0.813  & 0.157 \\
TRELLIS.2 & 19.1  \quad  & 0.905  & 0.062  \\
LHM     & 26.5  \quad  & 0.947  & 0.047\\
Ours & \textbf{28.3 } \quad  & \textbf{0.951}  &\textbf{ 0.043} \\
\bottomrule
\end{tabular}%
\end{table}
\begin{table}[h]
  \small
  \caption{Quantitative evaluation of inference time.}
  \label{Tab_runtime}
  \centering

\resizebox{0.48\textwidth}{!}{
\setlength{\tabcolsep}{7pt}
  \begin{tabular}{lccccc}
    \toprule
    Methods & HumanNorm & TADA & En3D & Ours \\
    \midrule
     Time & 2h+  & 1h+ & 30min+  & \textbf{5min}\\   
    \bottomrule
  \end{tabular}
  }
\end{table}

\subsection{Comparisons}

\noindent \textbf{Quantitative comparison}.
To assess the quality of generated human geometry and texture separately, we render normal maps and RGB images from 32 views to compute FID. As mentioned above, FID measures the visual similarity and distribution discrepancy between renderings of generated 3D content and real images. We then evaluate text-to-3D consistency by calculating the CLIP score between the prompts and rendered views of the generated 3D humans. We randomly select 50 prompts for text-guided generation and render observations from the corresponding generated results to compute the CLIP score. Tab.~\ref{Tab_nvs} shows that our method outperforms other methods on both FID and CLIP score, indicating superior overall human-generation quality. In addition, Tab.~\ref{Tab_runtime} reports the average inference time of methods that can generate both geometry and texture with arbitrary topology. This demonstrates the significant speed advantage of our method over SDS-based methods. For fair comparison, all tests are conducted on one A800 GPU, and we compare only methods capable of generating both human geometry and texture. The runtime of each part of our pipeline is Geometry (1min52s), Texture Prior (40s), and Texture (2min36s).

To evaluate the performance of our diffusion renderer, we compare novel-view synthesis results. Tab.~\ref{Tab_human} shows that our diffusion renderer significantly outperforms the baselines in terms of PSNR, SSIM, and LPIPS. It provides high consistency and quality while also enabling free-viewpoint rendering. Overall, our method efficiently achieves high-quality and consistent 3D human geometry and texture generation from text descriptions.

\begin{figure*}[!t]
  \centering
   \includegraphics[width=1.0\textwidth]{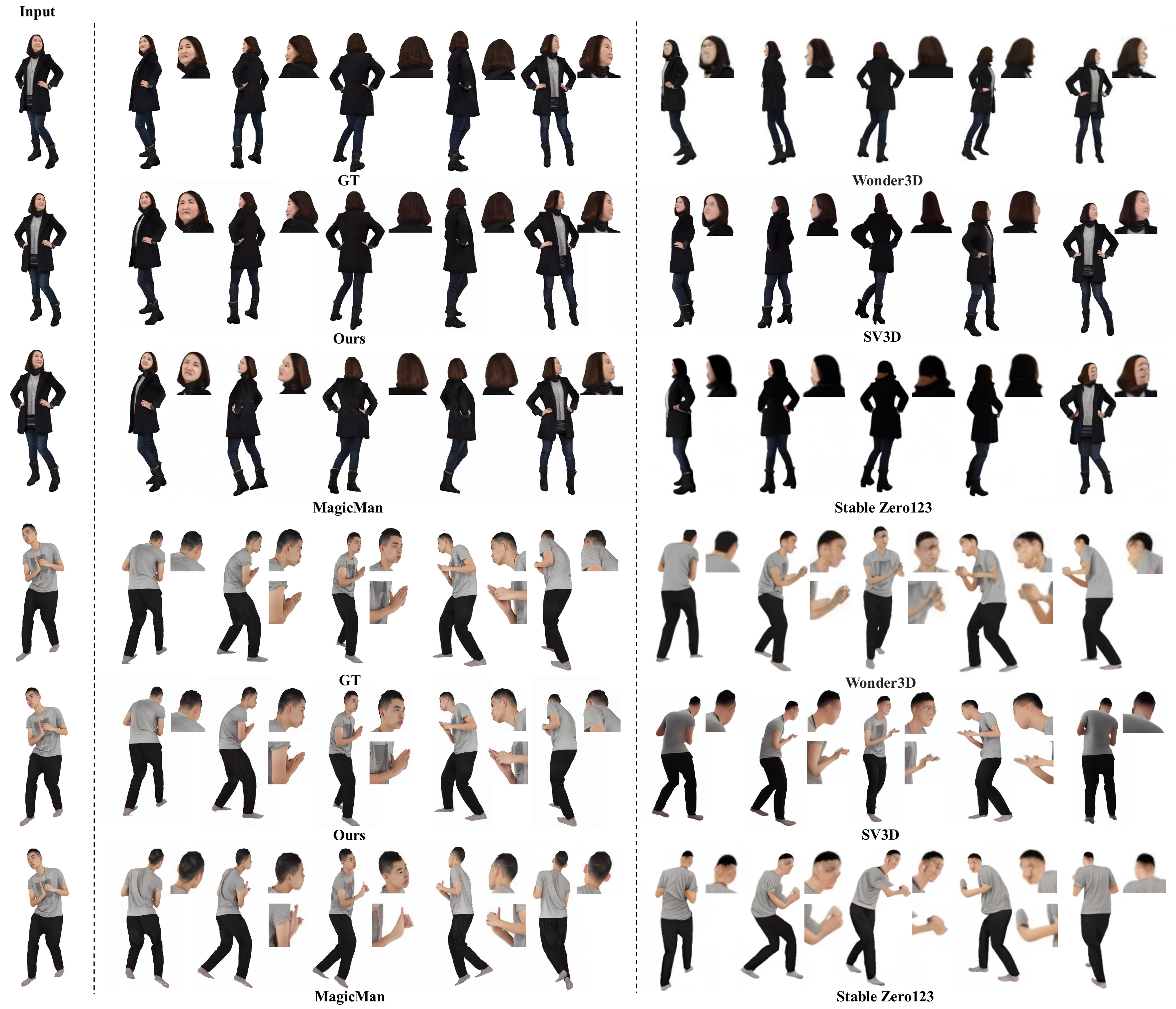}
   \caption{Qualitative comparison of novel-view synthesis.}
    \label{fig:compare_mvs}
\end{figure*}

\noindent \textbf{Qualitative comparison}.
Although our strategy uses geometry and texture priors derived from SMPL, as shown in Fig.~\ref{Figure:Teaser}, our method still supports the generation of humans with loose clothing. In the following, we qualitatively compare human-generation quality and novel-view synthesis quality. Fig.~\ref{fig:compare_texture} and Fig.~\ref{fig:compare_geometry} show that our method produces higher-quality and more realistic 3D humans. As shown in Fig.~\ref{fig:compare_mvs}, the diffusion renderer also generates more consistent and higher-quality novel views than prior methods~\cite{voleti2025sv3d,long2024wonder3d,he2024magicman,liu2023zero123}.

\begin{figure}[!h]
  \centering
   \includegraphics[width=1.0\linewidth]{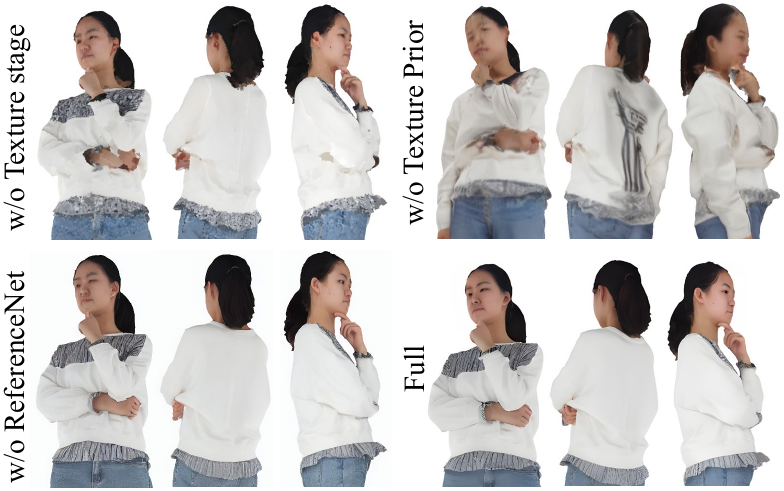}
   \vspace {-0.4cm} 
   \caption{Qualitative ablation of the modules in the texture stage.}
   \label{fig:ablation}
\end{figure}

\begin{table}[!t]
\centering
\caption{Ablation studies for the ReferenceNet and the Texture Prior in the diffusion renderer.}
\label{Table_Tex_Ablation}
\setlength{\tabcolsep}{9pt}
\begin{tabular}{lccc}
\toprule
 Methods & PSNR $\uparrow$ &  SSIM $\uparrow$ & LPIPS $\downarrow$ \\
\hline
w/o ReferenceNet & 24.6  \quad  & 0.897  & 0.053 \\
w/o Texture Prior & 19.3 \quad  & 0.729  & 0.071 \\
w/o Texture stage & 22.7 \quad  & 0.801  & 0.060 \\
Full & 28.3  \quad & 0.951  &0.043 \\
\bottomrule
\end{tabular}%
\end{table}

\subsection{Ablation study}
 In the geometry and texture-prior stages, removing any module would break the entire pipeline, making module-level ablation infeasible. Therefore, we focus only on certain hyperparameters in these two stages, and the corresponding ablation studies are provided in the supplementary material.
 
 In the texture stage, we compare the full diffusion renderer with variants without ReferenceNet or without the texture prior in Fig.~\ref{fig:ablation} and Tab.~\ref{Table_Tex_Ablation}. Both qualitative and quantitative ablations demonstrate the importance of each component. Removing the texture prior significantly degrades 3D consistency, highlighting the necessity of the texture-prior acquisition stage. In addition, without ReferenceNet or the full texture stage, the generated results become less realistic and less consistent. 
\subsection{User Study}
To better evaluate different methods qualitatively, we conduct a user study to analyze the quality of generated results, as detailed in the supplementary material.
\subsection{Application}
\begin{figure*}[!h]
		\centering
		\includegraphics[width=0.8\textwidth]{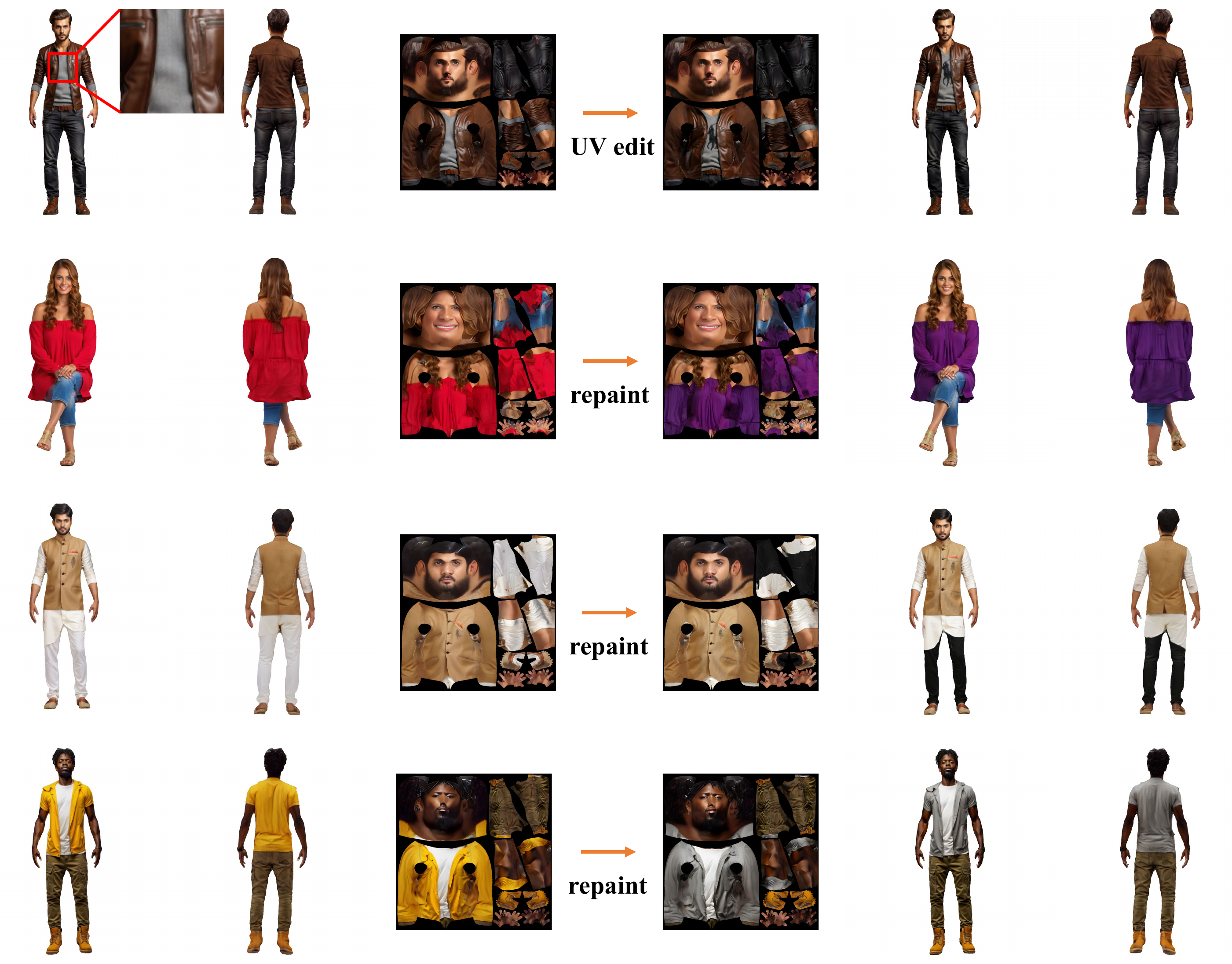}
		\vspace{-0.3cm}
		\caption{Local-region editing of 3D humans via SMPL UV repainting.}
		  \label{fig:tex_edit}
  \vspace{-0.5cm}
	\end{figure*}

\noindent \textbf{Texture editing}.
We enable flexible texture editing by modifying the SMPL UV and the front-view texture during the texture-prior stage, which allows local appearance editing without introducing artifacts into other regions, as demonstrated in Fig.~\ref{fig:tex_edit}.
\\
\noindent \textbf{3D human animation}. Since our model directly generates explicit digital assets (meshes and textures), it naturally supports skeleton-driven animation. Our animation pipeline transfers standard skeletal motion (from motion capture or keyframing) to the generated 3D human character. The main steps include auto-rigging, animation retargeting, and linear blend skinning (LBS). This pipeline handles complex poses well and can also support cases in which the character carries objects. We provide a demo video with more dynamic animation results and multi-view renderings in the accompanying material.

\section{Limitations and Discussions}

\subsection{Limitations and Failure Cases}

\begin{figure}[!h]
		\centering
		\includegraphics[width=0.5\textwidth]{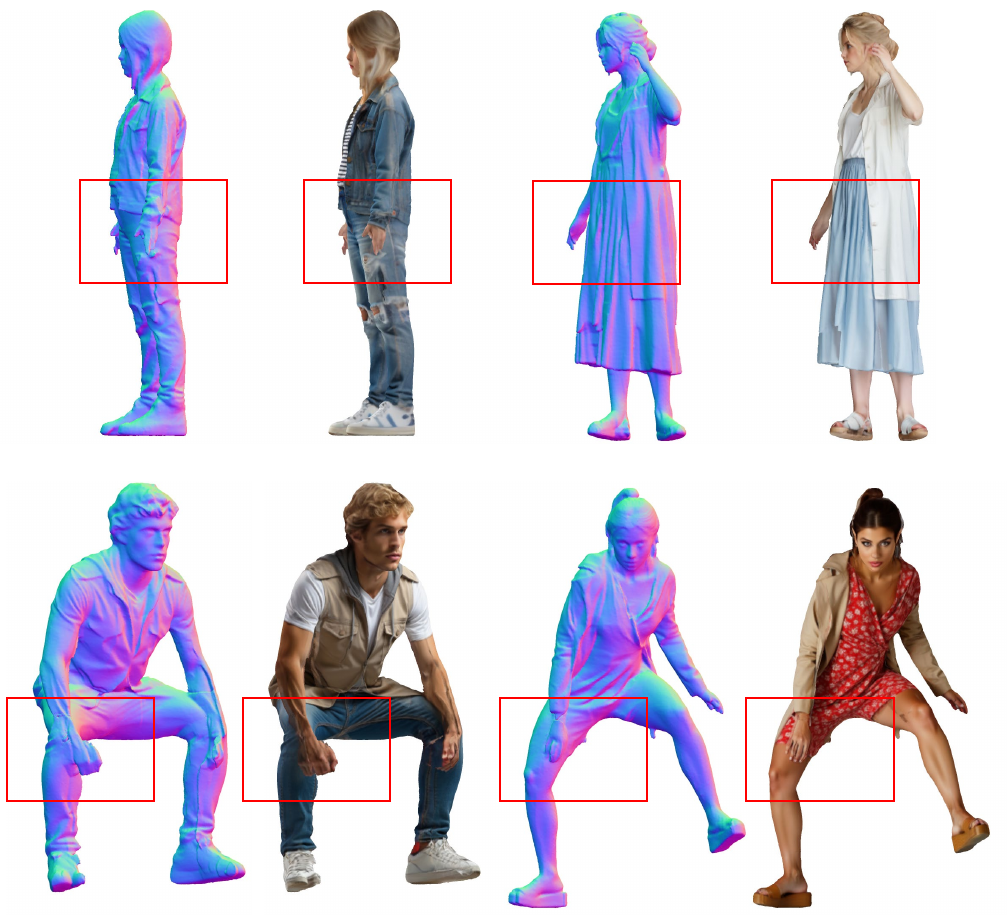}
		\vspace{-0.3cm}
		\caption{Failure cases involving fingers.}
		  \label{fig:failure_case_fingers}
  \vspace{-0.5cm}
	\end{figure}

\begin{figure}[!h]
		\centering
		\includegraphics[width=0.5\textwidth]{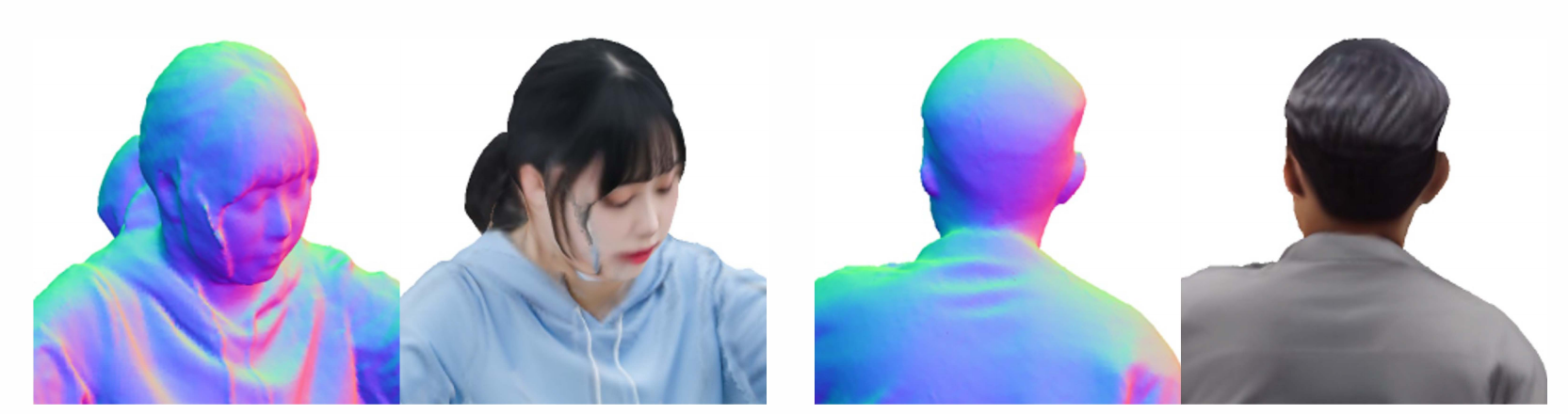}
		\vspace{-0.3cm}
		\caption{Failure cases involving hair.}
		  \label{fig:failure_case_hair}
  \vspace{-0.5cm}
	\end{figure}

\begin{figure}[!h]
		\centering
		\includegraphics[width=0.5\textwidth]{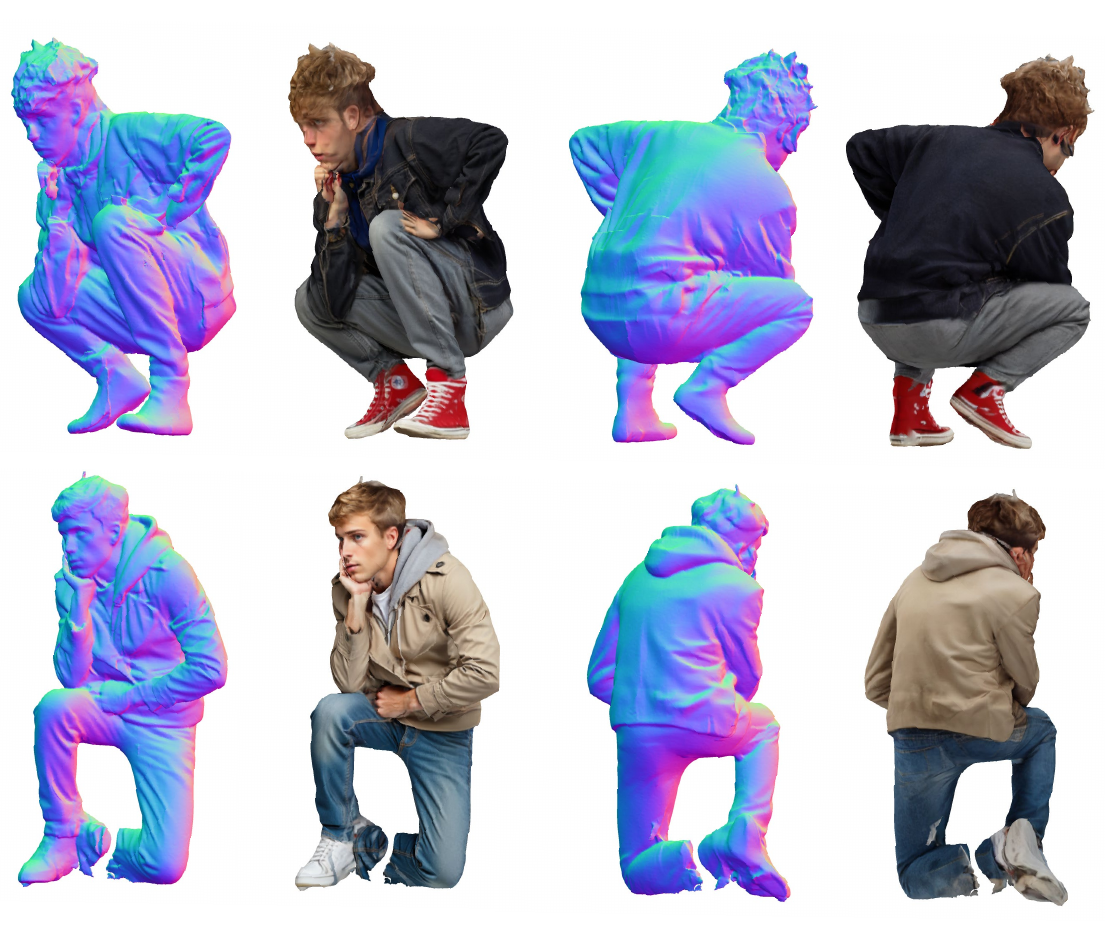}
		\vspace{-0.3cm}
		\caption{Failure cases of generated results under out-of-distribution pose guidance.}
		  \label{fig:failure_case_pose}
  \vspace{-0.5cm}
	\end{figure}

\noindent \textbf{Fine-detail degradation in small regions (fingers and hair)}. Since fingers and hair occupy relatively small image regions and exhibit high-frequency structures, as shown in Fig.~\ref{fig:failure_case_fingers} and Fig.~\ref{fig:failure_case_hair}, our diffusion-based generation may produce over-smoothed results or inconsistent fine details (e.g., fused fingers or blurry or unstable hair strands), especially under complex poses or severe self-occlusion.\\
\noindent \textbf{Out-of-distribution poses}. Our method may underperform when guided by poses that are rarely covered by the training distribution, particularly heavily occluded, unusual, or highly articulated poses. As demonstrated in Fig.~\ref{fig:failure_case_pose}, the pose guidance may be insufficient to resolve severe ambiguities, leading to local anatomical artifacts or implausible geometry.

\noindent \textbf{Inference latency}. To prioritize generation quality and controllability, our pipeline is not fully end-to-end and involves multiple stages and modules, which increases inference time compared with single-pass feed-forward approaches. 
\subsection{Robustness evaluation}
\begin{figure}[!h]
		\centering
		\includegraphics[width=0.5\textwidth]{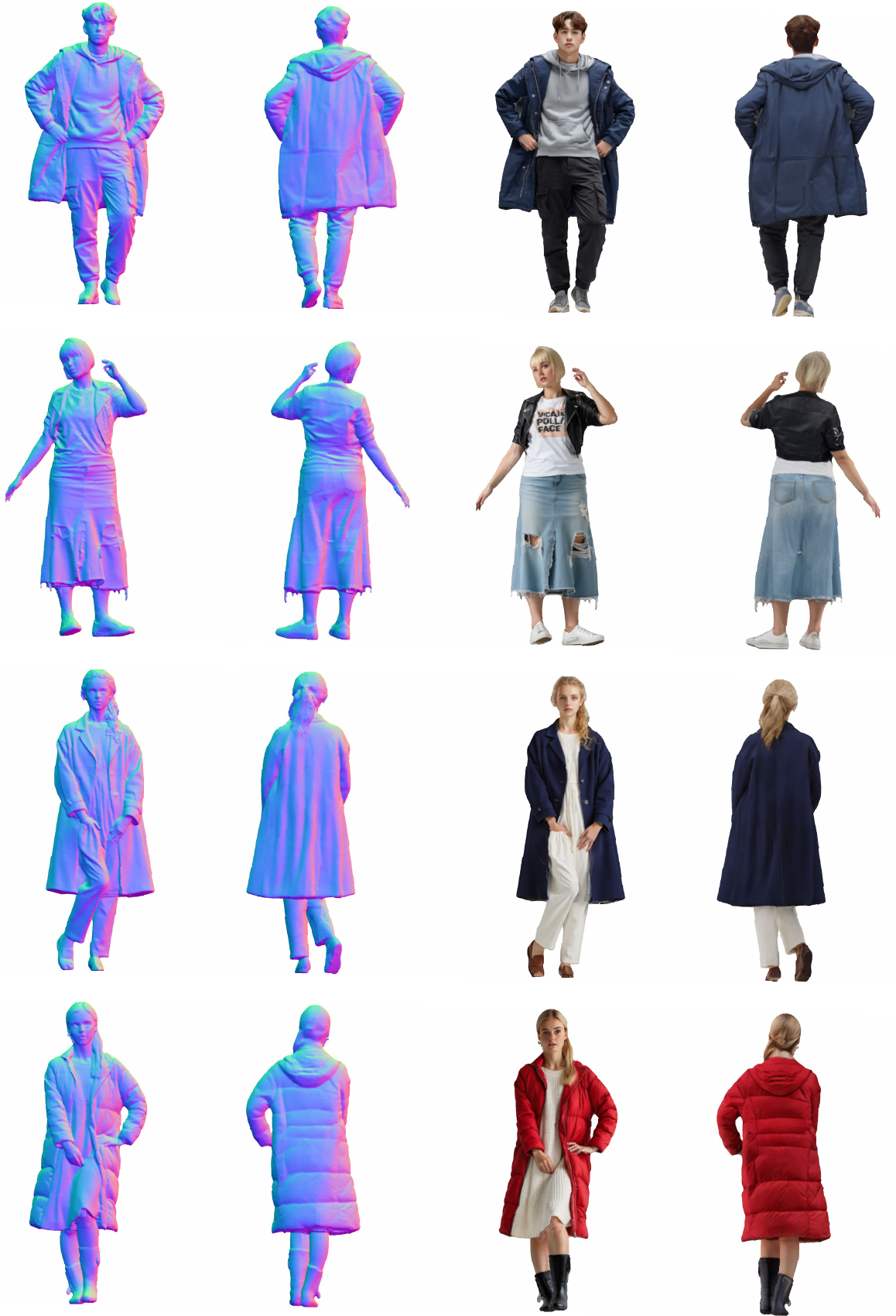}
		\vspace{-0.3cm}
		\caption{Generated results for prompts describing loose clothing.}
		  \label{fig:loose_cloth}
  \vspace{-0.5cm}
	\end{figure}

\noindent \textbf{Robustness to SMPL topology and loose clothing}. Unlike reconstruction methods, our approach takes a known 3D SMPL body as input and uses it only as initialization for normal-based optimization. Therefore, it is not affected by SMPL estimation noise or depth ambiguities typical of reconstruction settings. In practice, as illustrated in Fig.~\ref{fig:loose_cloth}, even for prompts describing very loose clothing, the generated body geometry remains stable and does not exhibit noticeable degradation.
\\

\begin{figure}[!h]
		\centering
		\includegraphics[width=0.5\textwidth]{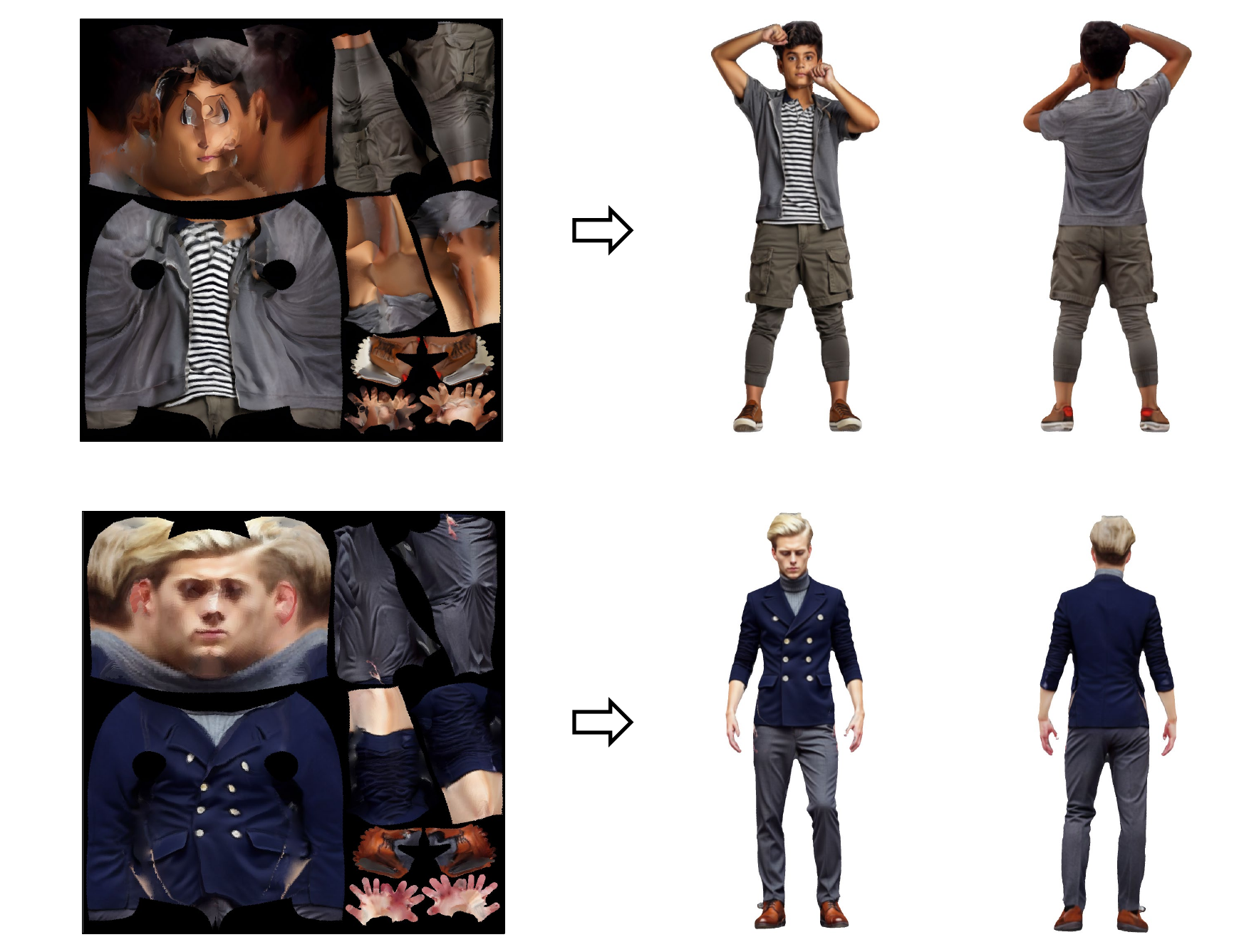}
		\vspace{-0.3cm}
		\caption{Generated results under suboptimal UV inpainting conditions.}
		  \label{fig:robustness_uv}
  \vspace{-0.5cm}
	\end{figure}

\noindent \textbf{Robustness to suboptimal UV inpainting results}. We assess the robustness of the diffusion renderer under imperfect intermediate UV inpainting outputs. As demonstrated in Fig.~\ref{fig:robustness_uv}, the results show the robustness of the diffusion renderer to such imperfections. 

\begin{figure}[!h]
		\centering
		\includegraphics[width=0.5\textwidth]{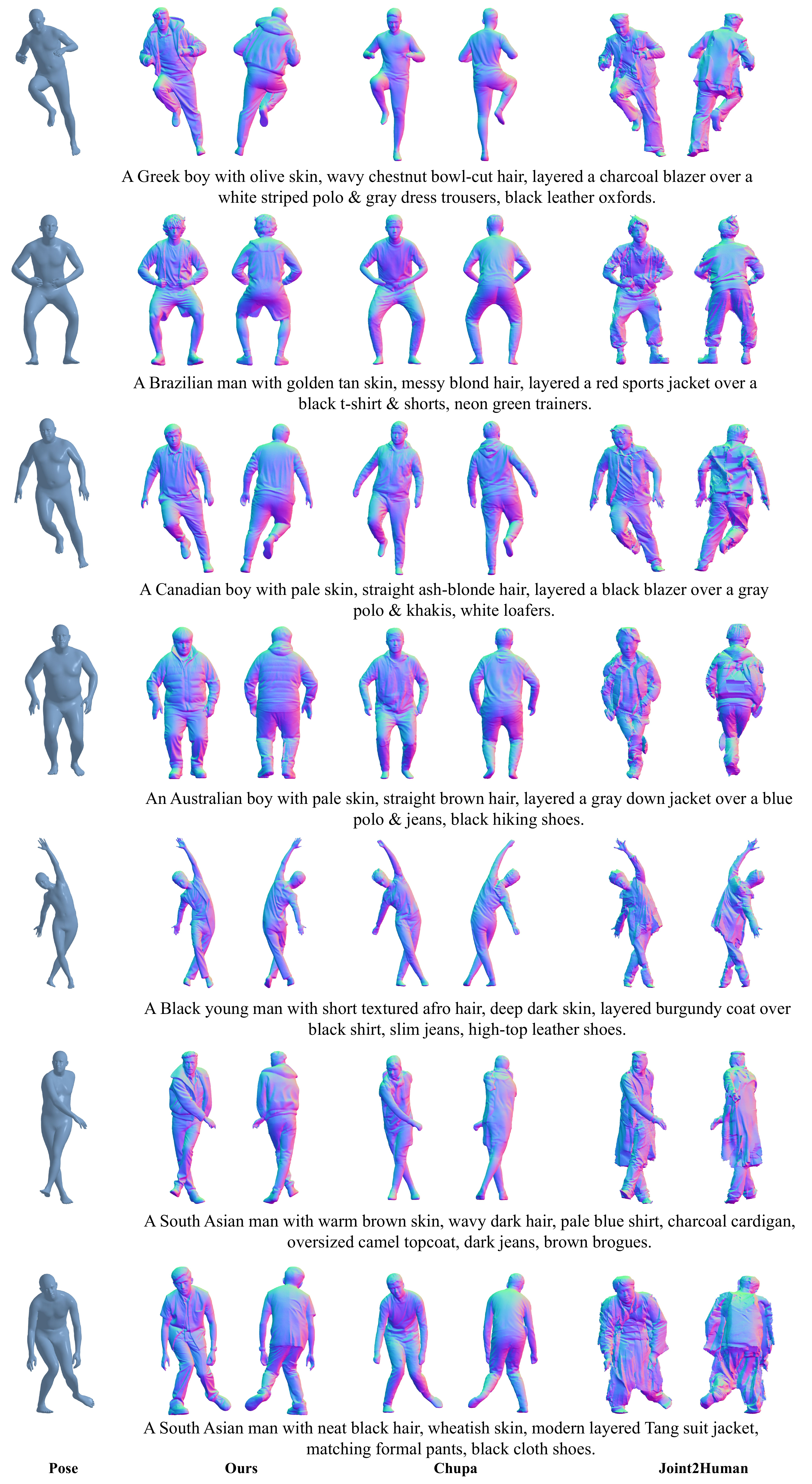}
		\vspace{-0.3cm}
		\caption{Qualitative comparison of human geometry under complex poses.}
		  \label{fig:geo_diff_pose}
  \vspace{-0.5cm}
	\end{figure}

\begin{figure}[!h]
		\centering
		\includegraphics[width=0.5\textwidth]{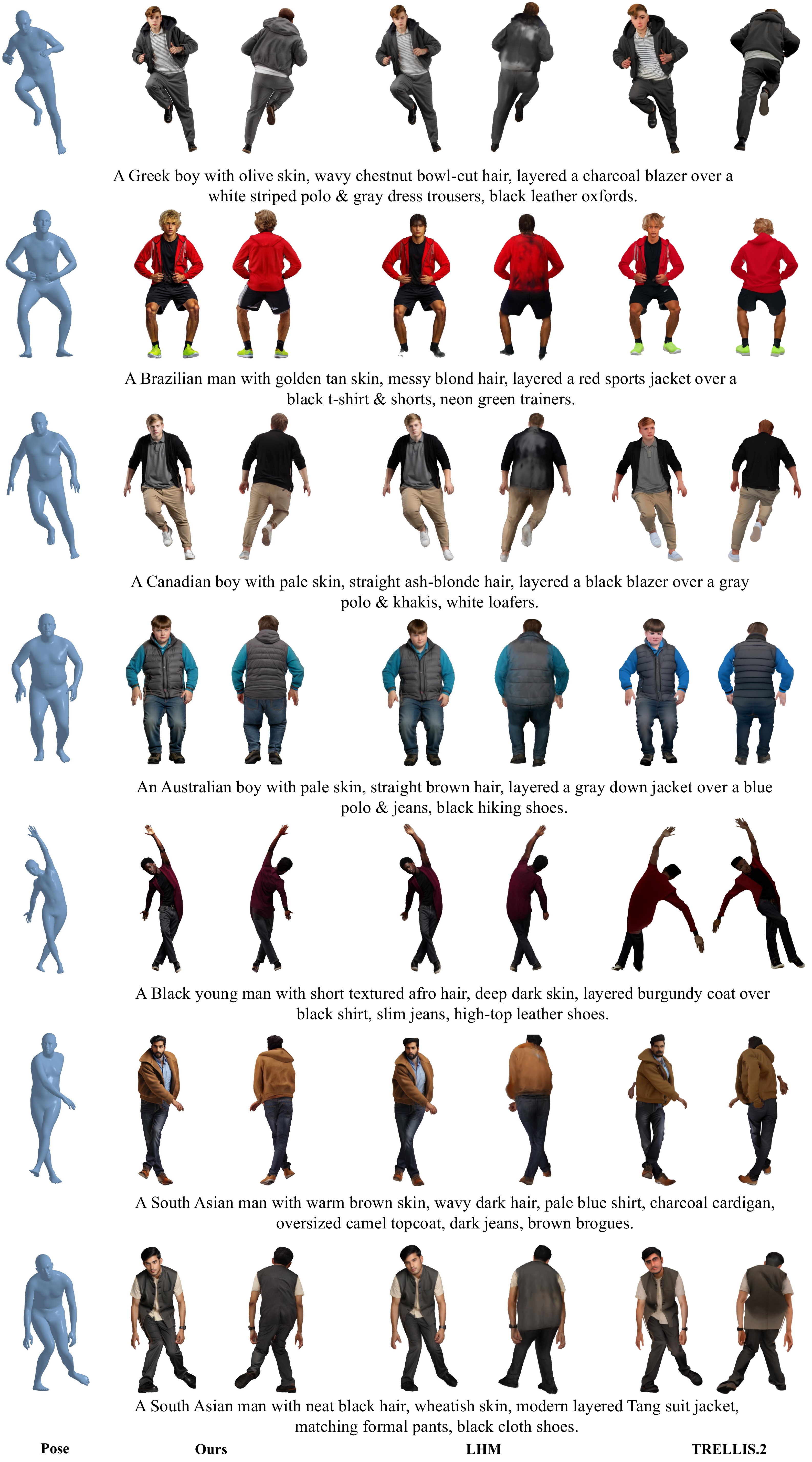}
		\vspace{-0.3cm}
		\caption{Qualitative comparison of human texture under complex poses.}
		  \label{fig:tex_diff_pose}
  \vspace{-0.5cm}
	\end{figure}

\noindent \textbf{Robustness to different poses}. As shown in Fig.~\ref{fig:geo_diff_pose} and Fig.~\ref{fig:tex_diff_pose}, thanks to extensive pretraining on large-scale synthetic data that better approximates the real-world human data distribution, our model demonstrates stronger robustness to pose variation than other methods.

\section{Conclusions }
In this work, we propose TGRHuman, a novel framework for 3D human generation. Unlike previous methods, we decouple geometry and texture generation through explicit 2D observation generation and supervision. Both the geometry and texture stages leverage consistent 2D multi-view observations and explicit optimization to achieve efficient and high-quality 3D human generation. In the geometry stage, we propose a high-resolution generative module for multi-view normals together with a geometry-carving strategy for human shape reconstruction. In the texture stage, we introduce a texture-prior acquisition strategy and a diffusion renderer for free-view rendering. Experimental results demonstrate that our model outperforms existing text-to-3D human generation methods in both geometry and texture quality. Although our current study focuses on text-to-3D human generation, the method is inherently flexible and can be extended to image-conditioned settings.
\\

\noindent \textbf{Acknowledgements.} 
This work was supported in part by National Key R\&D Program of China (2023YFC3082100) and Science Fund for Distinguished Young Scholars of Tianjin (No.22JCJQJC00040).





\clearpage
\bibliographystyle{elsarticle-num} 
\bibliography{example}






\end{document}